\documentclass{article}
\usepackage{ijcai22}

\usepackage{times}
\usepackage{soul}
\usepackage{url}
\usepackage[hidelinks]{hyperref}
\usepackage[utf8]{inputenc}
\usepackage[small]{caption}
\usepackage{graphicx}
\usepackage{amsmath}
\usepackage{amsthm}
\usepackage{amssymb}
\usepackage{booktabs}
\usepackage{algorithm}
\usepackage{algorithmic}
\usepackage{multirow}
\usepackage{color,xcolor}
\usepackage{comment}

\title{C3-STISR: Scene Text Image Super-resolution with Triple Clues}

\author{
	Minyi Zhao$^{1}$\thanks{This work is done while authors are interns in ByteDance.} \quad Miao Wang$^{2}$ \quad Fan Bai$^{1}$  \\{\bf Bingjia Li}$^1$\footnotemark[1] \quad {\bf Jie Wang}$^2$ \quad {\bf Shuigeng Zhou}$^1$\thanks{Corresponding author.} \\
	$^1$Shanghai Key Lab of Intelligent Information Processing,  and School of \\ Computer Science, Fudan University, Shanghai 200438, China\\
	$^2$ByteDance, China\\
	$^{1}$\{zhaomy20, fbai19, bjli20, sgzhou\}@fudan.edu.cn \\ $^{2}$\{wangmiao.01, wangjie.bernard\}@bytedance.com \\
}

\begin{document}
\maketitle

\begin{abstract}
 Scene text image super-resolution (STISR) has been regarded as an important pre-processing task for text recognition from low-resolution scene text images. Most recent approaches use the recognizer's feedback as clue to guide super-resolution. However, directly using recognition clue has two problems: 1) \emph{Compatibility}. It is in the form of probability distribution, has an obvious modal gap with STISR --- a pixel-level task; 2) \emph{Inaccuracy}. it usually contains wrong information, thus will mislead the main task and degrade super-resolution performance. %{\color{red}Inspired by the fact that human will use recognizable information and linguistical knowledge to guess the fuzzy text and think about the stroke details of each character to recover a text image. }
% To tackle these problems, i
In this paper, we present a novel method C3-STISR that jointly exploits the recognizer's feedback, visual and linguistical information as clues to guide super-resolution. Here, visual clue is from the images of texts predicted by the recognizer, which is informative and more compatible with the STISR task; while linguistical clue is generated by a pre-trained character-level language model, which is able to correct the predicted texts. We design effective extraction and fusion mechanisms for the triple cross-modal clues to generate a comprehensive and unified guidance for super-resolution. Extensive experiments on TextZoom show that C3-STISR outperforms the SOTA methods in fidelity and recognition performance. Code is available in \url{https://github.com/zhaominyiz/C3-STISR}.
\end{abstract}
% Besides, the proposed TRIGE can also boost text detection and video text tracking tasks.
\section{Introduction}

\emph{Scene text recognition} (STR), which aims to recognize texts from input scene images has wide applications such as auto-driving~\cite{zhang2020street} and scene-text-based image understanding~\cite{singh2019towards}. Although great progress has been made in STR due to the development of deep learning, recognition performance on low-resolution (LR) text images is still unsatisfactory. %{\color{red}Besides, these heavily deblurred LR text images also severely degrade the Quality of Experience.}
Ergo, \emph{scene text image super-resolution} (STISR)~\cite{wang2020scene} is gaining popularity as a pre-processing technique to recover the missing details in LR images for boosting text recognition performance. % {\color{red}and fidelity performance}.

\begin{figure}
	\begin{center}
		\includegraphics[width=0.9\linewidth]{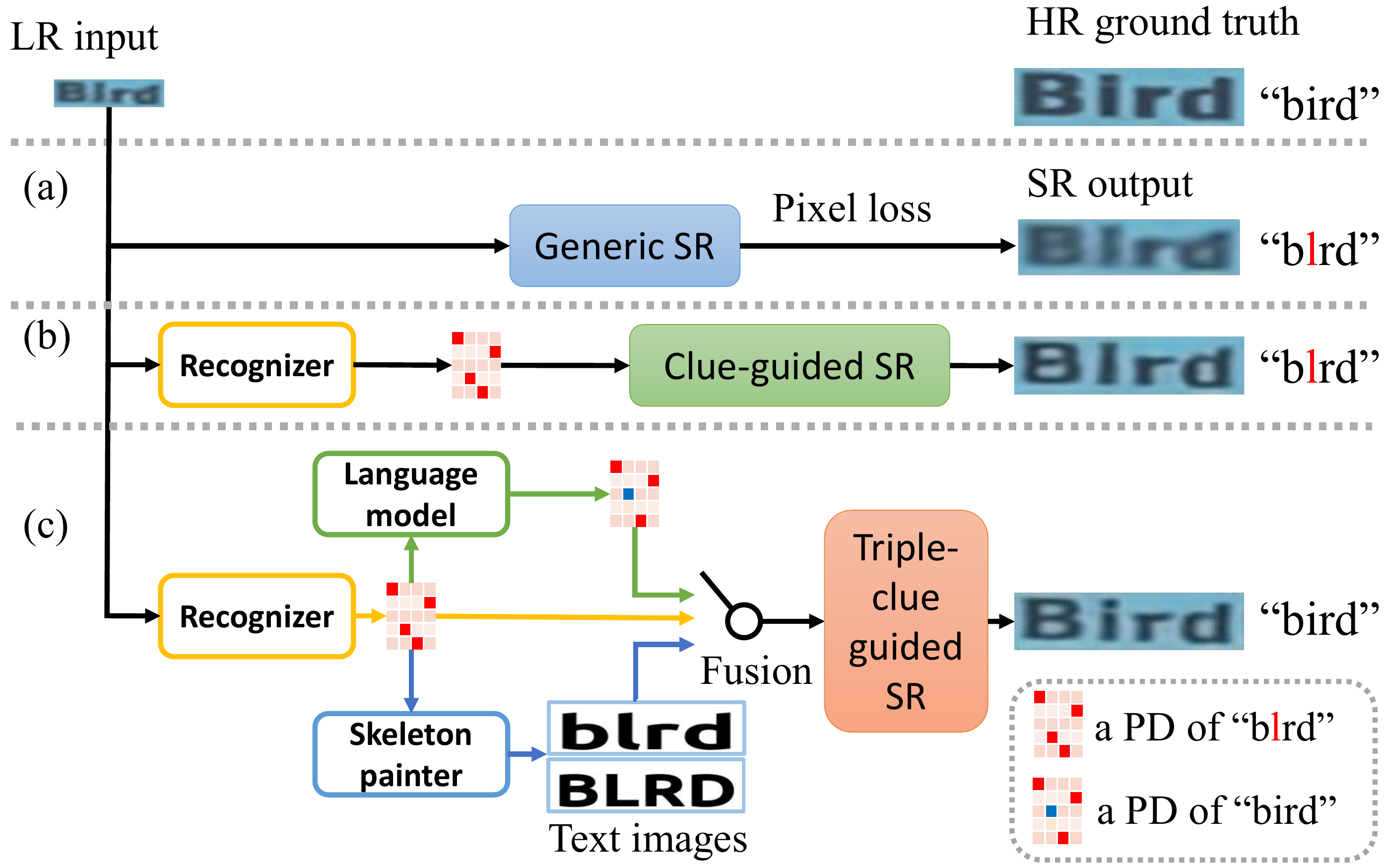}
	\end{center}
	\caption{Schematic illustration of existing STISR works roughly falling into two types: (a) generic methods, (b) clue-guided methods, and (c) our method C3-STISR that jointly exploits triple cross-modality clues: linguistical (up), recognition (middle), and visual (down), to boost super-resolution. PD: probability distribution.}
	\label{fig-intro}
\end{figure}

Existing STISR works roughly fall into two categories: generic high-resolution (HR) methods and clue-guided solutions. As shown in Fig.~\ref{fig-intro}, the generic methods~\cite{xu2017learning,pandey2018binary} usually try to learn missed details through HR-LR image pairs with pixel loss functions (\emph{e.g}. $L1$ or $L2$ loss). They treat text images as normal images and disregard their text-specific characteristics, usually cannot achieve satisfied recognition performance. Recently, more and more works attempt to take text-specific characteristics as clues to guide super-resolution, which leads to better performance in terms of image quality and recognition accuracy. For example, \cite{chen2021scene} takes the attention map and recognition result of the recognizer as clues to compute text-focused loss.
\cite{ma2021text} uses the recognition result as text-prior clue to iteratively conduct super-resolution. \cite{chen2021text} introduces stroke-level recognition clue to generate more distinguishable images.

Although these methods have definitely improved the recognition accuracy, their designs have some obvious shortcomings: 1) They mostly use the recognizer's feedback as clue to guide super-resolution, ignore other potentially useful information such as visual and linguistical information. 2) The widely used recognition clue is in the form of probability distribution~(PD), which has an obvious modal gap with STISR --- a low-level vision task, so there is a modal compatibility issue. 3) The recognizer's feedback is usually inaccurate (the recognition accuracy on LR/HR images is only 26.8\%/72.4\%, see Sec.~\ref{sec-sota}), thus will mislead the following super-resolution, especially in some tough scenarios. For example, in Fig.~\ref{fig-intro}(c), the recognizer's feedback is a PD of ``blrd'', but the ground truth is ``bird''. Such error in the feedback will inevitably impact super-resolution.

%To address the aforementioned issues,
Imagine how humans will repair LR text images in practice. In addition to the information directly from the images, they may also exploit character compositional/ structural information and linguistical knowledge to guess the blurred characters and words. With this in mind, in this paper we present a novel method C3-STISR that jointly exploits the recognizer's feedback, visual and linguistical information as clues to guide super-resolution, as shown in Fig.~\ref{fig-intro}(c). Concretely, the visual clue is extracted from the painted images of texts predicted by the recognizer, which is informative and more compatible with the STISR task, and thus will lead to better recovery (in Fig.~\ref{fig-intro}(c), a clearer and better `B' is gotten due to the usage of visual clue), while the linguistical clue is generated by a pre-trained character-level language model, which is able to correct the predicted text (in Fig.~\ref{fig-intro}(c), ``blrd'' is corrected to ``bird''). Furthermore, regarding that these clues are in different modalities, we first extract them in a divide-and-conquer way, and then aggregate them. We develop effective clue extractors and a unified gated fusion module that integrates the triple clues as a comprehensive guidance signal for super-resolution.

Main contributions of this paper are summarized as follows:
%\begin{itemize}
1) We propose a novel method C3-STISR to jointly utilize recognition, visual, and linguistical clues to guide super-resolution. Comparing with existing methods, C3-STISR can generate higher quality text images with the help of newly introduced visual and linguistical clues.
2) We design a powerful clue generator that extracts the triple cross-modal clues in a divide-and-conquer manner, and then fuse them to a comprehensive and unified one.
3) We conduct extensive experiments over the TextZoom dataset, which show that C3-STISR significantly outperforms the state-of-the-art approaches.
%\end{itemize}

\section{Related Work}
% Here we first review the related works that roughly fall into two groups: generic approaches and clue-guided approaches, according to whether they use text-specific clues, and then highlight the major differences between our method and existing clue-guided ones.

Here we review the related works that roughly fall into two groups: generic approaches and clue-guided approaches, according to whether they use text-specific clues.

%\subsection{Generic Approaches}
\textbf{Generic approaches}. These methods treat STISR as a general SR problem and recover LR images via pixel information captured by pixel loss functions. In particular, SRCNN~\cite{dong2015image} designs a three-layer convolutional neural network for the SR task. \cite{xu2017learning} and SRResNet~\cite{ledig2017photo} adopt generative adversarial networks to generate distinguishable images. \cite{pandey2018binary} combines convolutional layers, transposed convolution, and sub-pixel convolution layers to extract and upscale features. RCAN~\cite{zhang2018image} and SAN~\cite{dai2019second} introduce attention mechanisms to boost the recovery. Nevertheless, such approaches ignore text-specific characteristics, cannot achieve optimal performance.

%\subsection{Clue-guided Approaches}
\textbf{Clue-guided approaches}. Recent approaches focus on text-specific characteristics of the images and utilize them as clues to boost the recovery. They usually use an additional recognizer to conduct clue-guided super-resolution. Specifically, \cite{wang2019textsr,fang2021tsrgan,nakaune2021skeleton} calculate text-specific losses to enhance text recognition. \cite{wang2020scene} introduces TSRN and gradient profile loss to capture sequential and text-specific information of text images. PCAN~\cite{zhao2021scene} is proposed to learn sequence-dependent and high-frequency information of the reconstruction. STT~\cite{chen2021scene} makes use of character-level clue from a pre-trained transformer recognizer to conduct text-focused super-resolution. TPGSR~\cite{ma2021text} and \cite{ma2022text} extract predicted probability distribution or semantic feature as clues to recover low quality images. TG~\cite{chen2021text} uses stroke-level clue to generate more distinguishable images. Although these methods have definitely improved recognition accuracy, the clue from the recognizer is mainly in a probability distribution modality imcompatible with the STISR task, and usually inaccurate, which limits the improvement of recognition performance.

\section{Method}
Here we first give an overview of our method C3-STISR (meaning \emph{triple clues for STISR}), then present the triple-clue guided super-resolution backbone. Subsequently, we introduce the extraction and fusion components of the triple clues, followed by the design of loss function.

\begin{figure*}
	\begin{center}
		\includegraphics[width=0.82\linewidth]{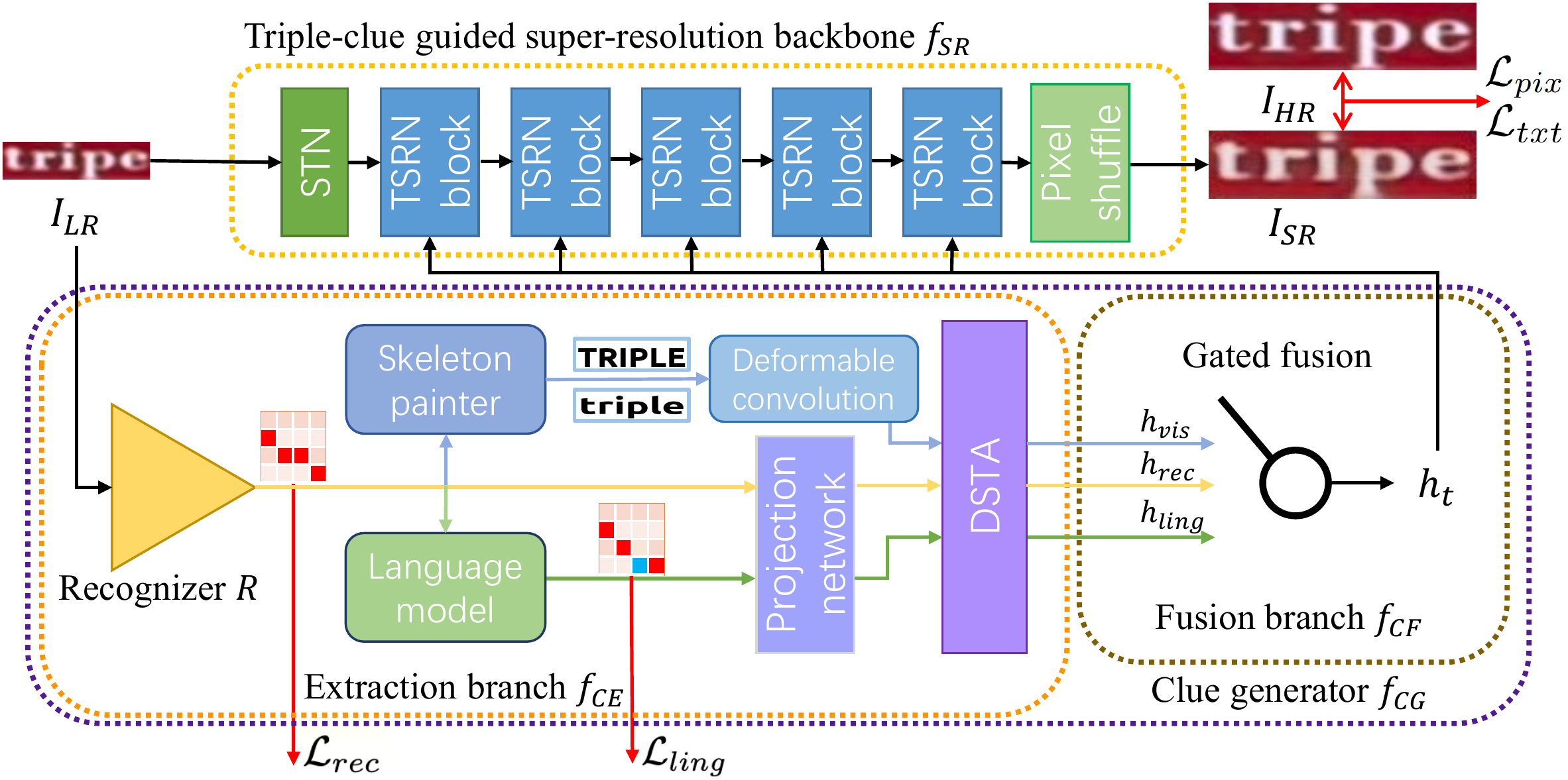}
	\end{center}
	\caption{The architecture of our method C3-STISR.}
	\label{fig-trige}
\end{figure*}

\subsection{Overview}
Given a low-resolution image $I_{LR}$ $\in$ $\mathbb{R}^{C \times N}$. Here, $C$ is the number of channels of each image, $N$ = $H \times W$ is the
collapsed spatial dimension, $H$ and $W$ are the height and width of image $I_{LR}$. Our aim is to produce a super-resolution (SR) image $I_{SR}$ $\in$ $\mathbb{R}^{C \times (4 \times N)}$ based on the input LR image $I_{LR}$ and some text-specific clue $h_t$.
Fig.~\ref{fig-trige} shows the architecture of our method C3-STISR, which is composed of two major components: the \emph{triple-clue guided super-resolution backbone} $f_{SR}$ that takes $I_{LR}$ and $h_t$ as input to generate a super-resolution image $I_{SR}=f_{SR}(I_{LR},h_t)$, and the \emph{clue generator} $f_{CG}$ that generates the clue $h_t$ to guide super-resolution. Specifically, $f_{CG}$ consists of two subcomponents: the \emph{clue extraction branch} $f_{CE}$ and the \emph{clue fusion branch} $f_{CF}$. The former generates the triple clues: recognition clue $h_{rec}$, visual clue $h_{vis}$ and linguistical clue $h_{ling}$ based on the feedback of a recognizer $R$ with $I_{LR}$ as input, \emph{i.e.}, \{$h_{rec},h_{vis},h_{ling}\}=f_{CE}(R(I_{LR}))$. Then, the latter fuses the triple clues to generate the comprehensive clue $h_t$ for super-resolution, i.e., $h_{t}=f_{CF}(h_{rec},h_{vis},h_{ling})$. During model training, the HR image $I_{HR}$ (ground truth) of each training LR image is taken as supervision to evaluate the pixel and text-specific losses.

\subsection{Triple-clue Guided Super-Resolution Backbone}
We design the backbone in the following way: 1) Notice that in the TextZoom dataset~\cite{wang2020scene}, the HR-LR pairs are manually cropped and matched by humans, which may incur several pixel-level offsets. Following previous works, the backbone starts with a Spatial Transformer Network (STN)~\cite{jaderberg2015spatial}. 2) Five modified TSRN blocks are employed to recover $I_{LR}$ with the guidance of $h_t$. The clue $h_t$ is concatenated with the feature map extracted by the convolution layers of TSRN blocks at channel dimension. 3) A pixel shuffle module is applied to reshaping the super-resolution image. 4) Two different losses $\mathcal{L}_{pix}$ and $\mathcal{L}_{txt}$ are used to provide pixel and text-specific supervision, respectively. In particular, the $L_2$ pixel loss ($\mathcal{L}_{pix}$) and the text-focused loss ($\mathcal{L}_{txt}$)~\cite{chen2021scene} are separately adopted to trade-off fidelity and recognition performance:
\begin{equation}
	\label{eq-pix}
	\mathcal{L}_{pix}=||I_{HR}-I_{SR}||_2,
\end{equation}
\begin{equation}
	\label{eq-per}
	\mathcal{L}_{txt}=\lambda_1 a||A_{HR}-A_{SR}||_1 + \lambda_2 KL(p_{SR},p_{HR}),
\end{equation}
where $A$ and $p$ are the attention map and probability distribution predicted by a fixed transformer-based recognizer, respectively. KL denotes the \textit{Kullback-Leibler divergence}, and $\lambda_1$ and $\lambda_2$ are two hyper-parameters.

\subsection{Clue Generator}\label{sec-cg}
The clue generator aims to generate a comprehensive clue $h_t$ to guide the super-resolution backbone. To this end, we first extract triple cross-modal clues: recognition clue $h_{rec}$, visual clue $h_{vis}$ and linguistical clue $h_{ling}$ in a divide-and-conquer manner. Then, we fuse them to output $h_t$. Now, we start with the introduction of the clue extraction branch.

\subsubsection{Clue Extraction Branch}\label{sec-ce}
Clue extraction can be divided into two steps: first extracting the initial cross-modal clues, and then transforming them into corresponding pixel-level ones for fusion. % and guide the super-resolution.
%(we mark as $\hat{h}_{rec}, \hat{h}_{vis}$, and $\hat{h}_{ling}$), and then process them into corresponding pixel-level one (\textit{i.e.,} $h_{rec}, h_{vis}$, and $h_{ling}$) for clue fusion.

\textbf{$h_{rec}$ extraction}.
The recognition clue $h_{rec}$ is computed from the probability distribution predicted by the recognizer $R$: $h_{rec}$=$f_{rec}(R(I_{LR}))$, and $R(I_{LR})$ $\in$ $\mathbb{R}^{L \times |\mathcal{A}|}$, $h_{rec}$ $\in$ $\mathbb{R}^{C' \times N}$. Here, $C'$, $L$ and $|\mathcal{A}|$ denote the channel number of hidden state, the max predicted length and the length of alphabet $\mathcal{A}$, respectively. $f_{rec} := \mathbb{R}^{L \times |\mathcal{A}|} \rightarrow \mathbb{R}^{C' \times N}$, is a processing network that transforms the probability distribution $R(I_{LR})$ to a pixel feature map and performs error reduction via masking uncertain information. Here, the processing network is implemented by a projection network and a deformable spatiotemporal attention (DSTA) block~\cite{zhao2021recursive}. In particular, the projection network consists of four transposed convolution layers followed by batch normalization and a bilinear interpolation; while the DSTA block utilizes the powerful deformable convolution~\cite{dai2017deformable} to compute a spatial attention map for masking uncertain information.
Considering that the performance of the recognizer can heavily influence $h_{rec}$, we adopt the distillation loss~\cite{ma2021text} to finetune the recognizer $R$:
\begin{equation}
	\label{eq-rec}
	\mathcal{L}_{rec}=k_1 ||R(I_{LR})-R(I_{HR})||_1 + k_2 KL(R(I_{LR}),R(I_{HR})),
\end{equation}
where $k_1,k_2$ are two hyper-parameters.

\textbf{$h_{vis}$ extraction}.
%{\color{red}
Given the predicted probability distribution $R(I_{LR})$, the goal of the visual clue extractor is to generate the visual information of the text image derived from the recognition result of $I_{LR}$. To this end, we first introduce a decoding function $f_{de}:= \mathbb{R}^{L \times |\mathcal{A}|} \rightarrow \mathbb{N}^{L}$ to decode the probability distribution to a text string, and then utilize a skeleton painter $f_{sp}:=\mathbb{N}^{L} \rightarrow \mathbb{R}^{C \times N}$ to draw the text image. The drawn text image presents the skeleton of the text to be recognized, and provides useful structural information for STISR. Here, we use \textit{Python Image Library} (PIL) as $f_{sp}$ to draw black-white text images. Nevertheless, the generated text image is in pixel level and has two shortcomings, which makes it fail to directly guide super-resolution. First, the prediction confidence is lost during decoding, which may exacerbate the propagation of errors. Second, the text image is generated in horizontal direction with fixed font, while the recognition clue is interpolated to the pixel level, which may incur motion and shape misalignment. Ergo, we also design a processing network $f_{vis}:= \mathbb{R}^{C \times N} \rightarrow \mathbb{R}^{C' \times N}$ to handle these problems. Specifically, $f_{vis}$ consists of a deformable convolution~\cite{dai2017deformable} that uses $h_{rec}$ to align and compensate the text image and a DSTA block for error reduction. Finally, $h_{vis}$ is extracted as follows:
	\begin{equation}
		\label{eq-vis}
		h_{vis}=f_{vis}(f_{sp}(f_{de}(R(I_{LR}))),h_{rec}).
	\end{equation}
%}

\textbf{$h_{ling}$ extraction}.
Given $R(I_{LR})$, the linguistical clue extractor is to correct $R(I_{LR})$ via a language model $f_{LM}$ and output the corrected probability distribution $p_{LM}$, i.e., $p_{LM}=f_{LM}(R(I_{LR}))$. To achieve this, we employ a pre-trained bidirectional cloze network~\cite{fang2021read} as the language model (LM) to perform character-level correction. The LM is first pre-trained via spelling mutation and recovery with a corpus~\cite{merity2016pointer}, and then finetuned via the distillation loss to adapt to the super-resolution task.
%learn the linguistical knowledge and to be adapted to the super-resolution task.
That is, we finetune the LM as follows:
	\begin{equation}
		\label{eq-ling}
		\mathcal{L}_{ling}=k_1 ||p_{LM}-R(I_{HR})||_1 + k_2 KL(p_{LM},R(I_{HR})).
	\end{equation}
	
We also design a processing network $f_{ling}:= \mathbb{R}^{L \times |\mathcal{A}|} \rightarrow \mathbb{R}^{C' \times N}$ for the linguistical clue. Similar to $f_{rec}$, $f_{ling}$ consists of a projection network and a DSTA block for error reduction as the correction operation may also be inaccurate.

\subsubsection{Clue Fusion Branch}\label{sec-cf}

With the clue extraction branch, the triple clues are transformed into unified pixel feature maps of $C' \times N$ size. Here, we employ a modified gated fusion~\cite{xu2021boosting} to fuse the clues softly.
Specifically, given the three pixel-level clues $h_{rec},h_{ling}$ and $h_{vis}$, we first adopt several dilated convolution layers to extract their features. Then, we stack these features with the LR image $I_{LR}$ in the channel dimension, and utilize a group of convolution layers to generate a mask $M \in \mathbb{R}^{3 \times C' \times N}$. After performing softmax along the first dimension of $M$, we get the fused clue $h_t$ as follows:
\begin{equation}
	\label{eq-fuse}
	h_t = M[0,:] \otimes {h}_{rec} \oplus M[1,:] \otimes {h}_{ling} \oplus M[2,:] \otimes {h}_{vis},
\end{equation}
where $\otimes$ and $\oplus$ indicate pixel multiplication and pixel addition, respectively.

\subsection{Overall Loss Function}
There are four types of loss functions used in our method: the first is a pixel loss (Eq.~(\ref{eq-pix})), the second is for recognition performance (Eq.~(\ref{eq-per})), the third is for finetuning the recognizer (Eq.~(\ref{eq-rec})), and the last is for finetuning the LM (Eq.~(\ref{eq-ling})). Thus, the overall loss function is
\begin{equation}
	\label{eq-final}
	\mathcal{L} = \alpha_1 \mathcal{L}_{pix} + \alpha_2 \mathcal{L}_{txt} + \alpha_3 \mathcal{L}_{rec} + \alpha_4 \mathcal{L}_{ling},
\end{equation}
where $\alpha_1,\alpha_2,\alpha_3,\alpha_4$ are four hyper-parameters.

\subsection{Multi-stage Training}
To exploit the triple clues of different modalities to the greatest extent, the training process of our method is split into three steps: first, we pre-train the LM via spelling mutation and recovery. Second, we pre-train the recognition clue and visual clue extraction modules. Finally, integrating the pretrained LM with the other modules,  we finetune the whole model. Such a training scheme can ensure that the model does not forget the pre-trained linguistic knowledge.

\section{Performance Evaluation}
In this section, we first introduce the dataset and metrics used in the experiments and the implementation details. Then we compare our method with the state-of-the-art approaches. Finally, we conduct extensive ablation studies to validate the design of our method.

\subsection{Dataset and Metrics}
The \textbf{TextZoom}~\cite{wang2020scene} dataset consists of 21,740 LR-HR text image pairs collected by lens zooming of the camera in real-world scenarios. The training set has 17,367 pairs, while the test set is divided into three settings based on the camera focal length, namely easy (1,619 samples), medium (1,411 samples) and hard (1,343 samples).

We utilize recognition accuracy to evaluate the recognition performance of the method. We remove all the punctuations and convert uppercase letters to lowercase letters for calculating recognition accuracy, by following the settings of previous works~\cite{chen2021scene}. In addition, we use Peak Signal-to-Noise Ratio (PSNR) and Structural Similarity Index Measure (SSIM) to evaluate fidelity.

\subsection{Implementation Details}
Our model is implemented in PyTorch1.8. All experiments are conducted on 8 NVIDIA
Tesla V100 GPUs with 32GB memory. The model is trained using Adam~\cite{kingma2014adam} optimizer with a learning rate of 0.001. The batch size is set to 48. The recognizer $R$ used in our method is CRNN~\cite{shi2016end}. The hyper-parameters in our method are set as follows: $\lambda_1=10$, $\lambda_2=0.0005$, $k_1=1.0$, $k_2=1.0$, $\alpha_1=20$, $\alpha_2=20$, $\alpha_3=1$, $\alpha_4=0.2$, $C'=32$, which are recommended in \cite{chen2021scene,ma2021text}. The font used by the skeleton painter is ubuntu bold. Two text images (one uppercase, one lowercase) are generated by the skeleton painter for each LR image. Our training and evaluation are based on the following protocol: save the averagedly best model during training with CRNN as the recognizer, and use this model to evaluate the
\begin{table*}[t]
	%\begin{subtable}
	\centering
	\scalebox{0.85}{
		\begin{tabular}{c||c c c c|c c c c|c c c c}
			\hline
			\multirow{2}*{Method}  &  \multicolumn{4}{c|}{CRNN \cite{shi2016end}} & \multicolumn{4}{c|}{MORAN \cite{luo2019moran}} & \multicolumn{4}{c}{ASTER \cite{shi2018aster}}\\
			% \cline{3-14}
			\cline{2-13}
			~  & Easy & Medium & Hard & Average & Easy & Medium & Hard & Average & Easy & Medium & Hard & Average  \\
			\hline
			\hline
			BICUBIC  & 36.4\% & 21.1\% & 21.1\% & 26.8\% & 60.6\% & 37.9\% & 30.8\% & 44.1\% & 67.4\% & 42.4\% & 31.2\% & 48.2\%  \\
			% \hline
			HR  & 76.4\% & 75.1\% & 64.6\% & 72.4\% & 91.2\% & 85.3\% & 74.2\% & 84.1\% & 94.2\% & 87.7\% & 76.2\% & 86.6\%  \\
			\hline
			\hline
			SRCNN   & 41.1\%  & 22.3\%  & 22.0\%  & 29.2\%  & 63.9\%  & 40.0\%  & 29.4\%  & 45.6\%  & 70.6\%  & 44.0\%  & 31.5\%  & 50.0\% \\
			% \hline
			SRResNet  & 45.2\% & 32.6\% & 25.5\% & 35.1\% & 66.0\% & 47.1\% & 33.4\% & 49.9\% & 69.4\% & 50.5\% & 35.7\% & 53.0\%    \\
			RCAN & 46.8\% & 27.9\% & 26.5\% & 34.5\% & 63.1\% & 42.9\% & 33.6\% & 47.5\% & 67.3\% & 46.6\% & 35.1\% & 50.7\%    \\
			SAN  & 50.1\% & 31.2\% & 28.1\% & 37.2\% & 65.6\% & 44.4\% & 35.2\% & 49.4\% & 68.1\% & 48.7\% & 36.2\% & 52.0\%    \\
			% \hline
			TSRN  & 52.5\%  & 38.2\%  & 31.4\%  & 41.4\%  & 70.1\%  & 55.3\%  & 37.9\%  & 55.4\%  & 75.1\%  & 56.3\%  & 40.1\%  & 58.3\% \\
			% \cite{nakaune2021skeleton} & BMVC21 & 56.5\%  & 42.3\%  & 32.5\%  & 44.5\%  & 72.9\%  & 55.9\%  & 40.6\%  & 57.5\%  & 77.3\%  & 59.6\%  & 42.7\%  & 60.9\% \\
			STT  & 59.6\% & 47.1\% & 35.3\% & 48.1\% & 74.1\% & 57.0\% & 40.8\% & 58.4\% & 75.7\% & 59.9\% & 41.6\% & 60.1\%  \\
			PCAN & 59.6\% & 45.4\% & 34.8\% & 47.4\% & 73.7\% & 57.6\% & 41.0\% & 58.5\% & 77.5\% & 60.7\% & 43.1\% & 61.5\%  \\
			% \hline
			TG  & 61.2\% & 47.6\% & 35.5\% & 48.9\% & \textbf{75.8}\% & 57.8\% & 41.4\% & 59.4\%  & 77.9\% & 60.2\% & 42.4\% & 61.3\%  \\
			\hline
			\hline
			Baseline~(w/o clue) & 54.8\% & 42.9\% & 32.7\% & 44.2\% & 67.5\% & 52.7\% & 37.1\% & 53.4\%  & 72.3\% & 56.1\% & 38.5\% & 56.8\%  \\
			Ours (C3-STISR) & \textbf{65.2}\% & \textbf{53.6}\% & \textbf{39.8}\% & \textbf{53.7}\% & 74.2\% & \textbf{61.0}\% & \textbf{43.2}\% & \textbf{60.5}\%  & \textbf{79.1}\% & \textbf{63.3}\% & \textbf{46.8}\% & \textbf{64.1}\%  \\
			% \hline
			% \hline
			% TPGSR^{*} & Arxiv & 63.1\% & 52.0\% & 38.6\% & 51.8\% & 74.9\% & 60.5\% & 44.1\% & 60.5\%  & 78.9\% & 62.7\% & 44.5\% & 62.8\%  \\
			% Ours^{*} & - & 63.1\% & 52.0\% & 38.6\% & 51.8\% & 74.9\% & 60.5\% & 44.1\% & 60.5\%  & 78.9\% & 62.7\% & 44.5\% & 62.8\%  \\
			\hline
	\end{tabular}}
	\caption{Performance (recognition accuracy) comparison on TextZoom.}
	\label{tab:final result}
\end{table*}
other recognizers (MORAN, ASTER) and the three settings (Easy, Medium, Hard). %, which is more reasonable.

\subsection{Comparing with the SOTA Approaches}
\label{sec-sota}
Here we evaluate our method on \textbf{TextZoom}, and compare it with existing super-resolution models on three recognition models, including CRNN~\cite{shi2016end}, MORAN~\cite{luo2019moran} and ASTER~\cite{shi2018aster}. The results are presented in Tab.~\ref{tab:final result}. We can see that our method significantly improves the recognition accuracy. Taking CRNN as an example, comparing with the state-of-the-art method TG~\cite{chen2021text} that boosts the performance from 48.1\% to 48.9\% (increasing 0.8\%), our method lifts the accuracy from 48.9\% to 53.7\% (increasing 4.8\%). This demonstrates the effectiveness and advantage of our method.

\begin{table}[t]
%\begin{subtable}
\centering
\scalebox{0.95}{
\begin{tabular}{c|ccc}
\hline
\multirow{2}*{Method} & \multicolumn{3}{c}{Metric}\\
\cline{2-4}
~& PSNR & SSIM~($\times 10^{-2}$) & Avg Acc  \\
 \hline
 BICUBIC & 20.35 & 69.61 & 26.8  \\
 TSRN & 21.42 & 76.91 & 41.4  \\ %rec only
  STT & 21.05 & 76.14 & 48.1  \\ % ling only
  PCAN & \textbf{21.49} & \textbf{77.53} & 47.4  \\ % vis only
TG & 21.40 & 74.56 & \textbf{48.9}  \\ %rec+ling
Ours (C3-STISR)  & \textbf{21.51} & \textbf{77.21} & \textbf{53.7}  \\ % rec+vis
\hline
\end{tabular}}
\caption{Fidelity and recognition performance comparison with major existing methods.  The results are obtained by averaging that of three settings (Easy, Medium and Hard).}
\label{tab:perception}
\end{table}

 \begin{figure}
	\begin{center}
		\includegraphics[width=0.9\linewidth]{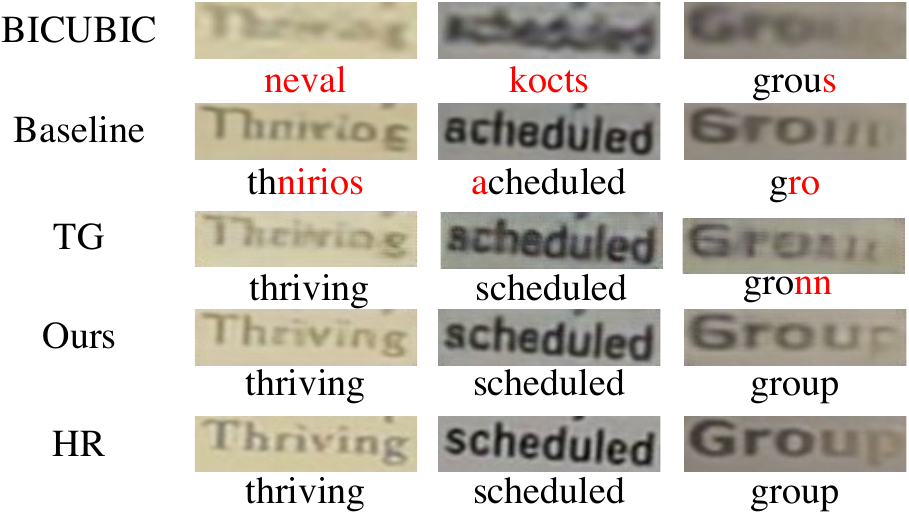}
	\end{center}
	\caption{Examples of generated SR images and recognition results from the SR images by different methods. Red characters are incorrectly recognized, and black characters are correctly recognized.}
	\label{fig-vis}
\end{figure}

We also present the results of fidelity (PSNR and SSIM) comparison with major existing methods in Tab.~\ref{tab:perception}. Our method is advantageous over or comparable to the SOTA in fidelity, while significantly outperforms the others in recognition performance. Furthermore, we visualize some examples in Fig.~\ref{fig-vis}. Compared with the other methods, C3-STISR can recover the blurry pixels better.
Experimental results on more recognizers, benchmarks, inference time-cost, and comparison with TPGSR are given in the supplementary material.
% We also discuss the inference speed in the supplementary material.

\subsection{Ablation Study}
Here, we conduct extensive ablation studies to validate the design of our method. The recognition performance is measured by the average accuracy of CRNN.

\begin{table}[t]
	%\begin{subtable}
	\centering
	\scalebox{0.95}{
		\begin{tabular}{c||ccc}
			\hline
			\multirow{2}*{Variant} & \multicolumn{3}{c}{Metric}\\
			\cline{2-4}
			~& PSNR & SSIM~($\times 10^{-2}$) & Avg Acc  \\
			\hline
			\hline
			w/o ft & 21.09 & 75.48 & 50.3 \\
			with ft & \textbf{21.14} & \textbf{75.98} & \textbf{52.2}  \\
			\hline
			\hline
			w/o compensation & 20.80 & 74.25 & 49.4\\
			with compensation & \textbf{21.21} & \textbf{76.38} & \textbf{51.7}  \\ % rec+vis
			\hline
			\hline
			w/o pt & \textbf{21.07} & 75.37 & 49.3\\ % Retrain
			w/o ft & 20.84 & \textbf{76.06} & 50.4\\
			with pt \& ft  & 20.94 & 75.78 & \textbf{51.0}  \\ % rec+vis
			\hline
	\end{tabular}}
	\caption{Ablation study on the design of the clue extraction branch. Here, ``ft'' and ``pt'' denote finetuning and pre-training, respectively.}
	\label{tab:ab:extraction}
\end{table}

\subsubsection{Design of Clue Extraction Branch}
%{\color{red}
We verify our design of the clue extraction modules. For simplicity, we check each clue separately. Results are in Tab.~\ref{tab:ab:extraction}.

\textbf{Recognition clue extraction}. The recognition clue is very important as it determines the other two types of clues: both linguistical clue and visual clue are extracted on the basis of the recognition clue. Ergo, we improve the recognition clue via finetuning. The first part (Rows 3-4) in Tab.~\ref{tab:ab:extraction} presents the results of without/with finetuning. We can see that without finetuning, the performance is degraded.

\textbf{Visual clue extraction}.
In our method, we employ the recognition clue to compensate and align the visual clue. We do this for two reasons: 1) the visual clue is generated from the drawn skeleton of the predicted text, which neglects the confidence from the recognizer. When the recognition result is uncertain, this exacerbates the propagation of error. 2) The black-white text image is generated in horizontal direction and using fixed font for convenience. That is, there is a modal gap (motion and shape misalignment) between the visual clue and the other two (recognition and linguistical) clues that are interpolated from probability distribution. Ergo, we utilize deformable convolutions to align them. For comparison, we also implement a variant that does not use compensation. As can be seen in the second part (Rows 5-6) of Tab.~\ref{tab:ab:extraction}, our design with compensation significantly boosts fidelity and recognition performance.

\textbf{Linguistical clue extraction}.
In C3-STISR, we apply pre-training and distillation loss $\mathcal{L}_{ling}$ (Eq.~(\ref{eq-ling})) to boost the knowledge learning from and the adaption to the linguistical domain. To check the effect of our design, we provide the performance of the variants that do not use pre-training or $\mathcal{L}_{ling}$. As shown in the third part (Rows 7-9 ) of Tab.~\ref{tab:ab:extraction}, such variants are inferior to that using both pre-training and $\mathcal{L}_{ling}$ in recognition accuracy.

\subsubsection{Design of Clue Fusion Branch}
There are many techniques to fuse multiple signals (\textit{e.g.} multi-head attention and deformable fusion~\cite{zhao2021recursive}). In our method, we fuse three clues via a modified gated fusion. The reason for our design lies in that after the projection network and deformable convolutions,
there is no more modal gap. Ergo, taking aligned clues as input, simple gated fusion is enough to fuse the triple clues via aggregating the pixels that are considered being correct. The experimental results are presented in Tab.~\ref{tab:ab:fusion}, from which we can see that the proposed gated fusion achieves the best performance among all the three fusion techniques.
\begin{table}[t]
%\begin{subtable}
\centering
\scalebox{0.95}{
\begin{tabular}{c|ccc}
\hline
\multirow{2}*{Fusion method} & \multicolumn{3}{c}{Metric}\\
\cline{2-4}
~& PSNR & SSIM~($\times 10^{-2}$) & Avg Acc  \\
 \hline
multi-head attention & 21.39 & 76.61 & 51.3 \\
DCN &21.31 &76.79 &51.5  \\
Gated fusion  & \textbf{21.51} & \textbf{77.21} & \textbf{53.7}  \\ % rec+vis
\hline
\end{tabular}}
\caption{Ablation study on the design of clue fusion branch.}
\label{tab:ab:fusion}
\end{table}
\begin{table}[t]
	%\begin{subtable}
	\centering
	\scalebox{0.85}{
		\begin{tabular}{c c c||ccc}
			\hline
			\multicolumn{3}{c||}{Clue} & \multicolumn{3}{c}{Metric}\\
			% \cline{3-9}
			\hline
			$h_{rec}$ & $h_{ling}$ & $h_{vis}$ & PSNR & SSIM~($\times 10^{-2}$) & Avg Acc  \\
			\hline
			\hline
			- &- &- & 21.38 & 76.82 & 44.2  \\
			\hline
			\hline
			$\checkmark$ &- &- & 21.14 & 75.98 & \textbf{52.2}  \\ %rec only
			- &$\checkmark$ &- & 20.94 & 75.78 & 51.0  \\ % ling only
			- & - &$\checkmark$ & \textbf{21.21} & \textbf{76.38} & 51.7  \\ % vis only
			\hline
			\hline
			$\checkmark$ & $\checkmark$ &- & 21.28 & \textbf{77.40} & \textbf{53.7}  \\ %rec+ling
			$\checkmark$ & - &$\checkmark$ & \textbf{21.38} & 77.39 & 53.5  \\ % rec+vis
			- & $\checkmark$ &$\checkmark$ & 21.31 & 76.57 & 52.9  \\ %ling+vis
			\hline
			\hline
			$\checkmark$ & $\checkmark$ &$\checkmark$ & \textbf{21.51} & 77.21 & \textbf{53.7}  \\ %all in
			\hline
			
			\hline
	\end{tabular}}
	\caption{Performance results of different combinations of 3 clues.}
	\label{tab:ab:comb}
\end{table}
\subsubsection{Different Combinations of the Triple Clues}
Above, we demonstrate the effectiveness of our designs through extensive experiments. Here, we check the performance of different combinations of the triple clues. The results are shown in Tab.~\ref{tab:ab:comb}. The baseline without any clues tends to repair each pixel in the image, which leads to good fidelity but low recognition accuracy. When clues are applied, the recognition accuracy is obviously improved. Among them, the recognition clue achieves the best recognition performance, and the visual clue outperforms the others in fidelity. The linguistical clue is inferior to the other two clues since STISR is a vision task. When two clues are combined, recognition-linguistical achieves the best accuracy. What is more, when recognition clue combines with visual clue, the fidelity is better than that of either single clue. This shows the effectiveness of the linguistical and visual clues. 
Finally, the combination of all the triple clues achieves the best performance in both fidelity (PSNR) and recognition performance, which shows that the proposed triple clues are complementary and all are required for better performance.

\begin{table}[t]
%\begin{subtable}
\centering
\scalebox{0.95}{
\begin{tabular}{c|ccc}
\hline
\multirow{2}*{Method} & \multicolumn{3}{c}{Metric}\\
\cline{2-4}
~& PSNR & SSIM~($\times 10^{-2}$) & Avg Acc  \\
 \hline
w/o MST & 19.84 & 74.31 & 51.1\\
w/o DSTA & 21.24 & 76.23 & 51.7 \\
Ours (C3-STISR)  & \textbf{21.51} & \textbf{77.21} & \textbf{53.7}  \\ % rec+vis
\hline
\end{tabular}}
\caption{Ablation study on multi-stage training (MST) and DSTA.}
\label{tab:ab:multi-DSTA}
\end{table}

\subsubsection{Effect of Multi-stage Training}
To exploit the potential of each clue to the greatest extent, we design a multi-stage training procedure. To check the effect of multi-stage training scheme, we compare the performance with and without the scheme. As shown in Tab.~\ref{tab:ab:multi-DSTA}, without the proposed multi-stage training, performance is degraded.

\subsubsection{Effect of DSTA}
As described above, we stack three DSTA~\cite{zhao2021recursive} blocks in our clue extraction branch to mask uncertain information. To check the effect of such design, we present the results without stacking DSTA blocks in Tab.~\ref{tab:ab:multi-DSTA}. Obviously, without DSTA, the performance is degraded, which demonstrates the effect of DSTA.

\begin{table}[t]
	\centering
	\scalebox{0.95}{
		\begin{tabular}{c|ccccc}
			\hline
			\multirow{2}*{~} & \multicolumn{5}{c}{$\alpha_4$}\\
			\cline{2-6}
			~&0.0 & 0.2 & 0.5 & 0.8& 1.0  \\
			\hline
			Avg Acc & 50.4 & \textbf{51.0} & 50.5 & 50.2 &50.7\\
			\hline
	\end{tabular}}
	\caption{The determination of $\alpha_4$. Here, we use only the linguistical clue as guidance signal.}
	\label{tab:ab:alpha4}
\end{table}
\subsubsection{Hyper-parameter Study}
We have some hyper-parameters to balance different losses. Here, $\lambda_1,\lambda_2$ are set as recommended in \cite{chen2021scene}, while $k_1,k_2,\alpha_1,\alpha_2,\alpha_3$ are set as suggested in \cite{ma2021text}. The remaining hyper-parameter to set is $\alpha_4$, which controls the language model. Here, we set $\alpha_4$ to relatively small values, aiming at retaining the linguistic knowledge as much as possible. We use gird search to determine $\alpha_4$. As shown in Tab.~\ref{tab:ab:alpha4}, when $\alpha_4=0.2$, the best performance is achieved. Ergo, $\alpha_4$ is set to 0.2 in our experiments.

\section{Conclusion}
In this paper, we present a novel method called C3-STISR that jointly utilizes recognition, visual, and linguistical clues to guide super-resolution. Comparing with the recognition clue used in existing works, the proposed visual clue is informative and more compatible, and the linguistical clue is able to correct error information in the recognition feedback. We develop an effective clue generator that first generates the triple cross-modal clues in a divide-and-conquer manner, and then aggregates them. Extensive experiments demonstrate the effectiveness and superiority of the proposed method.

\section*{Acknowledgments}
The work was supported in part by a ByteDance Research Collaboration Project.

 \newpage
\bibliographystyle{named}
\bibliography{ijcai22}

\begin{thebibliography}{}

\bibitem[\protect\citeauthoryear{Chen \bgroup \em et al.\egroup
  }{2021a}]{chen2021scene}
Jingye Chen, Bin Li, and Xiangyang Xue.
\newblock Scene text telescope: Text-focused scene image super-resolution.
\newblock In {\em CVPR}, pages 12026--12035, 2021.

\bibitem[\protect\citeauthoryear{Chen \bgroup \em et al.\egroup
  }{2021b}]{chen2021text}
Jingye Chen, Haiyang Yu, Jianqi Ma, Bin Li, and Xiangyang Xue.
\newblock Text gestalt: Stroke-aware scene text image super-resolution.
\newblock {\em arXiv preprint arXiv:2112.08171}, 2021.

\bibitem[\protect\citeauthoryear{Dai \bgroup \em et al.\egroup
  }{2017}]{dai2017deformable}
Jifeng Dai, Haozhi Qi, Yuwen Xiong, Yi~Li, Guodong Zhang, Han Hu, and Yichen
  Wei.
\newblock Deformable convolutional networks.
\newblock In {\em ICCV}, pages 764--773, 2017.

\bibitem[\protect\citeauthoryear{Dai \bgroup \em et al.\egroup
  }{2019}]{dai2019second}
Tao Dai, Jianrui Cai, Yongbing Zhang, Shu-Tao Xia, and Lei Zhang.
\newblock Second-order attention network for single image super-resolution.
\newblock In {\em CVPR}, pages 11065--11074, 2019.

\bibitem[\protect\citeauthoryear{Dong \bgroup \em et al.\egroup
  }{2015}]{dong2015image}
Chao Dong, Chen~Change Loy, Kaiming He, and Xiaoou Tang.
\newblock Image super-resolution using deep convolutional networks.
\newblock {\em TPAMI}, 38(2):295--307, 2015.

\bibitem[\protect\citeauthoryear{Fang \bgroup \em et al.\egroup
  }{2021a}]{fang2021tsrgan}
Chuantao Fang, Yu~Zhu, Lei Liao, and Xiaofeng Ling.
\newblock Tsrgan: Real-world text image super-resolution based on adversarial
  learning and triplet attention.
\newblock {\em Neurocomputing}, 455:88--96, 2021.

\bibitem[\protect\citeauthoryear{Fang \bgroup \em et al.\egroup
  }{2021b}]{fang2021read}
Shancheng Fang, Hongtao Xie, Yuxin Wang, Zhendong Mao, and Yongdong Zhang.
\newblock Read like humans: Autonomous, bidirectional and iterative language
  modeling for scene text recognition.
\newblock In {\em CVPR}, pages 7098--7107, 2021.

\bibitem[\protect\citeauthoryear{Jaderberg \bgroup \em et al.\egroup
  }{2015}]{jaderberg2015spatial}
Max Jaderberg, Karen Simonyan, Andrew Zisserman, et~al.
\newblock Spatial transformer networks.
\newblock In {\em NeurIPS}, pages 2017--2025, 2015.

\bibitem[\protect\citeauthoryear{Kingma and Ba}{2014}]{kingma2014adam}
Diederik~P Kingma and Jimmy Ba.
\newblock Adam: A method for stochastic optimization.
\newblock {\em arXiv preprint arXiv:1412.6980}, 2014.

\bibitem[\protect\citeauthoryear{Ledig \bgroup \em et al.\egroup
  }{2017}]{ledig2017photo}
Christian Ledig, Lucas Theis, Ferenc Husz{\'a}r, Jose Caballero, Andrew
  Cunningham, Alejandro Acosta, Andrew Aitken, Alykhan Tejani, Johannes Totz,
  Zehan Wang, et~al.
\newblock Photo-realistic single image super-resolution using a generative
  adversarial network.
\newblock In {\em CVPR}, pages 4681--4690, 2017.

\bibitem[\protect\citeauthoryear{Luo \bgroup \em et al.\egroup
  }{2019}]{luo2019moran}
Canjie Luo, Lianwen Jin, and Zenghui Sun.
\newblock Moran: A multi-object rectified attention network for scene text
  recognition.
\newblock {\em PR}, 90:109--118, 2019.

\bibitem[\protect\citeauthoryear{Ma \bgroup \em et al.\egroup
  }{2021}]{ma2021text}
Jianqi Ma, Shi Guo, and Lei Zhang.
\newblock Text prior guided scene text image super-resolution.
\newblock {\em arXiv preprint arXiv:2106.15368}, 2021.

\bibitem[\protect\citeauthoryear{Ma \bgroup \em et al.\egroup
  }{2022}]{ma2022text}
Jianqi Ma, Zhetong Liang, and Lei Zhang.
\newblock A text attention network for spatial deformation robust scene text
  image super-resolution.
\newblock {\em arXiv preprint arXiv:2203.09388}, 2022.

\bibitem[\protect\citeauthoryear{Merity \bgroup \em et al.\egroup
  }{2016}]{merity2016pointer}
Stephen Merity, Caiming Xiong, James Bradbury, and Richard Socher.
\newblock Pointer sentinel mixture models.
\newblock {\em arXiv preprint arXiv:1609.07843}, 2016.

\bibitem[\protect\citeauthoryear{Nakaune \bgroup \em et al.\egroup
  }{2021}]{nakaune2021skeleton}
Shimon Nakaune, Satoshi Iizuka, and Kazuhiro Fukui.
\newblock Skeleton-aware text image super-resolution.
\newblock 2021.

\bibitem[\protect\citeauthoryear{Pandey \bgroup \em et al.\egroup
  }{2018}]{pandey2018binary}
Ram~Krishna Pandey, K~Vignesh, AG~Ramakrishnan, et~al.
\newblock Binary document image super resolution for improved readability and
  ocr performance.
\newblock {\em arXiv preprint arXiv:1812.02475}, 2018.

\bibitem[\protect\citeauthoryear{Shi \bgroup \em et al.\egroup
  }{2016}]{shi2016end}
Baoguang Shi, Xiang Bai, and Cong Yao.
\newblock An end-to-end trainable neural network for image-based sequence
  recognition and its application to scene text recognition.
\newblock {\em TPAMI}, 39(11):2298--2304, 2016.

\bibitem[\protect\citeauthoryear{Shi \bgroup \em et al.\egroup
  }{2018}]{shi2018aster}
Baoguang Shi, Mingkun Yang, Xinggang Wang, Pengyuan Lyu, Cong Yao, and Xiang
  Bai.
\newblock Aster: An attentional scene text recognizer with flexible
  rectification.
\newblock {\em TPAMI}, 41(9):2035--2048, 2018.

\bibitem[\protect\citeauthoryear{Singh \bgroup \em et al.\egroup
  }{2019}]{singh2019towards}
Amanpreet Singh, Vivek Natarajan, Meet Shah, Yu~Jiang, Xinlei Chen, Dhruv
  Batra, Devi Parikh, and Marcus Rohrbach.
\newblock Towards vqa models that can read.
\newblock In {\em CVPR}, pages 8317--8326, 2019.

\bibitem[\protect\citeauthoryear{Wang \bgroup \em et al.\egroup
  }{2019}]{wang2019textsr}
Wenjia Wang, Enze Xie, Peize Sun, Wenhai Wang, Lixun Tian, Chunhua Shen, and
  Ping Luo.
\newblock Textsr: Content-aware text super-resolution guided by recognition.
\newblock {\em arXiv preprint arXiv:1909.07113}, 2019.

\bibitem[\protect\citeauthoryear{Wang \bgroup \em et al.\egroup
  }{2020}]{wang2020scene}
Wenjia Wang, Enze Xie, Xuebo Liu, Wenhai Wang, Ding Liang, Chunhua Shen, and
  Xiang Bai.
\newblock Scene text image super-resolution in the wild.
\newblock In {\em ECCV}, pages 650--666. Springer, 2020.

\bibitem[\protect\citeauthoryear{Xu \bgroup \em et al.\egroup
  }{2017}]{xu2017learning}
Xiangyu Xu, Deqing Sun, Jinshan Pan, Yujin Zhang, Hanspeter Pfister, and
  Ming-Hsuan Yang.
\newblock Learning to super-resolve blurry face and text images.
\newblock In {\em ICCV}, pages 251--260, 2017.

\bibitem[\protect\citeauthoryear{Xu \bgroup \em et al.\egroup
  }{2021}]{xu2021boosting}
Yi~Xu, Minyi Zhao, Jing Liu, Xinjian Zhang, Longwen Gao, Shuigeng Zhou, and
  Huyang Sun.
\newblock Boosting the performance of video compression artifact reduction with
  reference frame proposals and frequency domain information.
\newblock In {\em CVPRW}, pages 213--222, 2021.

\bibitem[\protect\citeauthoryear{Zhang \bgroup \em et al.\egroup
  }{2018}]{zhang2018image}
Yulun Zhang, Kunpeng Li, Kai Li, Lichen Wang, Bineng Zhong, and Yun Fu.
\newblock Image super-resolution using very deep residual channel attention
  networks.
\newblock In {\em ECCV}, pages 286--301, 2018.

\bibitem[\protect\citeauthoryear{Zhang \bgroup \em et al.\egroup
  }{2020}]{zhang2020street}
Chongsheng Zhang, Weiping Ding, Guowen Peng, Feifei Fu, and Wei Wang.
\newblock Street view text recognition with deep learning for urban scene
  understanding in intelligent transportation systems.
\newblock {\em IEEE Transactions on Intelligent Transportation Systems}, 2020.

\bibitem[\protect\citeauthoryear{Zhao \bgroup \em et al.\egroup
  }{2021a}]{zhao2021scene}
Cairong Zhao, Shuyang Feng, Brian~Nlong Zhao, Zhijun Ding, Jun Wu, Fumin Shen,
  and Heng~Tao Shen.
\newblock Scene text image super-resolution via parallelly contextual attention
  network.
\newblock In {\em MM}, pages 2908--2917, 2021.

\bibitem[\protect\citeauthoryear{Zhao \bgroup \em et al.\egroup
  }{2021b}]{zhao2021recursive}
Minyi Zhao, Yi~Xu, and Shuigeng Zhou.
\newblock Recursive fusion and deformable spatiotemporal attention for video
  compression artifact reduction.
\newblock In {\em MM}, pages 5646--5654, 2021.

\end{thebibliography}

\end{document}